\documentclass[a4paper,conference]{IEEEtran}
\ifCLASSINFOpdf
\else
\fi
\usepackage{graphicx}

\usepackage{balance}

\begin{document}
%
\title{Occluded Joints Recovery in 3D Human Pose Estimation based on Distance Matrix }

\author{\IEEEauthorblockN{Xiang Guo}
\IEEEauthorblockA{College of Engineering and Computer Science\\
The Australian National University}
\and
\IEEEauthorblockN{Yuchao Dai }
\IEEEauthorblockA{School of Electronics and Information \\
Northwestern Polytechnical University}
}


%


\maketitle
\begin{abstract}
Albeit the recent progress in single image 3D human pose estimation due to the convolutional neural network, it is still challenging to handle real scenarios such as highly occluded scenes. In this paper, we propose to address the problem of single image 3D human pose estimation with occluded measurements by exploiting the Euclidean distance matrix (EDM). Specifically, we present two approaches based on EDM, which could effectively handle occluded joints in 2D images. The first approach is based on 2D-to-2D distance matrix regression achieved by a simple CNN architecture. The second approach is based on sparse coding along with a learned over-complete dictionary. Experiments on the Human3.6M dataset show the excellent performance of these two approaches in recovering occluded observations and demonstrate the improvements in accuracy for 3D human pose estimation with occluded joints. \footnote{Yuchao Dai is the corresponding author (daiyuchao@gmail.com).}
\end{abstract}


%
\IEEEpeerreviewmaketitle

\section{Introduction}
\begin{figure}[!t]
\centering
\includegraphics[width=3.5in]{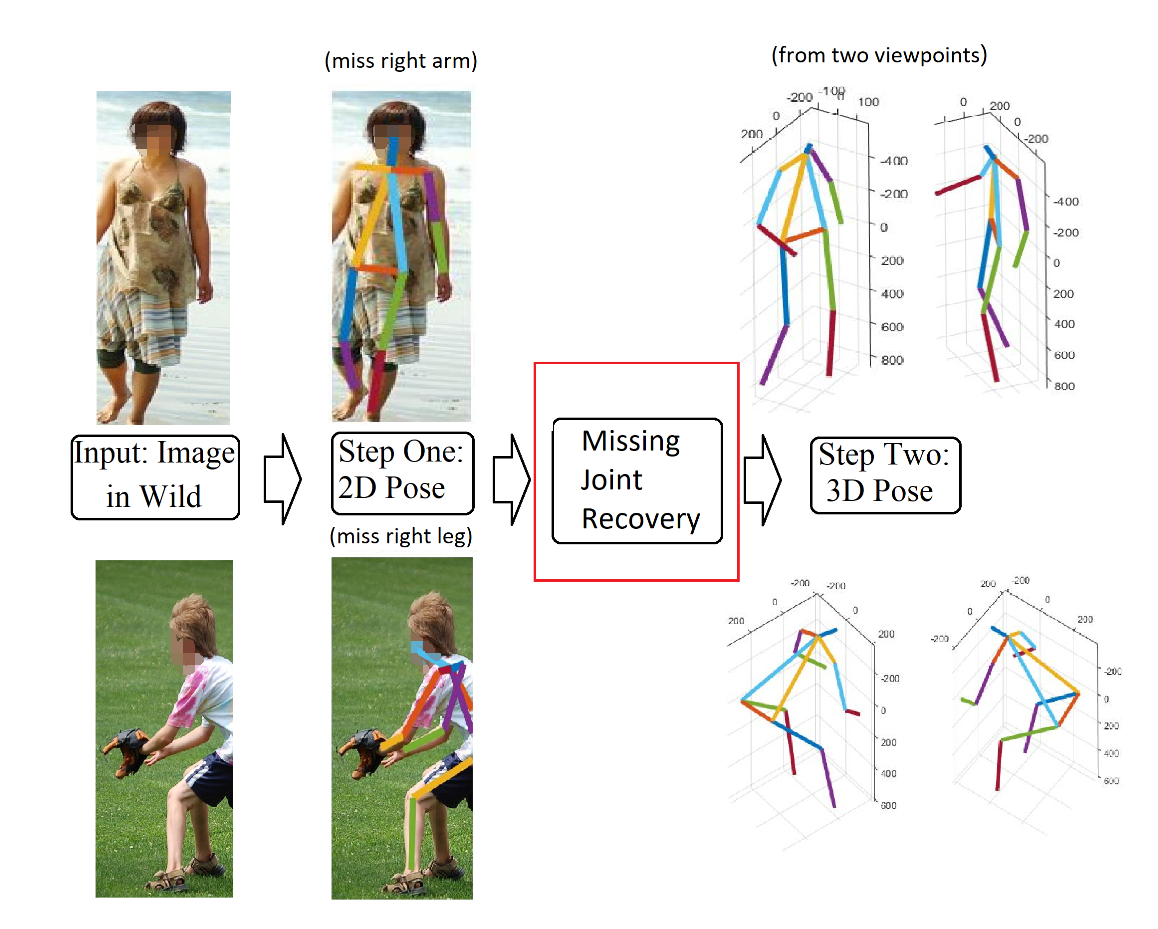}
\caption{Illustration of the two-step pipeline for handling occluded joints in 3D human pose estimation. Given occluded 2D human pose estimation as input, our approach successfully recovers the complete 3D human pose.}
\label{showcase}
\end{figure}

This paper is concerned with the commonly adopted pipeline in estimating 3D human pose from a single image \cite{DisMatrix,BayesianBased,DualSource,Sparse3D}, which generally consists of two consecutive steps: 1) use 2D joints detectors to localize joints in 2D image space and 2) estimate 3D pose from these observations by regressors learned from motion capture datasets. However, there is an inherent issue with this pipeline when applying in the uncontrolled real world scenarios. For images captured in the wild, there is a high possibility that human's bodies are occluded by each other or by other objects. The occluded joints could significantly affect the performance of 3D pose estimation in the second stage \cite{DisMatrix,BayesianBased,DualSource} (see Fig.~\ref{showcase} for an example). This practical and challenging problem has not been properly addressed \cite{BayesianBased,DualSource} \cite{DisMatrix,Sparse3D}.

In this paper, we proposed to tackle the above mentioned occluded joints problem by resorting to the Euclidean distance matrix, which is an efficient representation for pose in both 2D and 3D. Specifically, we are interested in recovering the occluded joints in 2D space and integrating it with 3D pose estimation pipeline. In order to examine the performance of our 2D occluded joints recovery approaches in the pipeline, we choose a state-of-the-art distance matrix regression based 3D human pose estimation method \cite{DisMatrix} as baseline.

In the general topic of occlusion handling, there have been various methods that could be explored to handle occluded joints. Sparse coding or dictionary based solutions recovers the occluded/missing measurements by a sparse linear combination of samples from the dictionary \cite{relatedwork:oimage}. Furthermore, many kinds of regressors have been built from motion capture datasets to enable the mapping from occluded pose to complete pose \cite{relatedwork:occldete}. The occluded limbs could also be restricted by introducing kinematic and orientation constraints \cite{relatedwork:selfoc}.

Instead of representing $N$ joints of one pose in the Cartesian space by a $N\times 2$ vector in 2D space or a $N\times 3$ vector in 3D space, we use a $N\times N$ Euclidean distance matrix (EDM), which consists of the Euclidean distance between each pair of joints. EDM owns three benefits 1) EDM is invariant to the image rotation and translation; 2) EDM is a way of encoding $N$ separate joints and obtaining the correlation between each pair of joints which benefits occluded joints recovery and 3) EDM based system is embeddable to our pipeline.

Based on EDM, we have developed two approached which achieve remarkable performance in recovering occluded joints. The first one is a CNN regressor. We build a fully convolutional neural network to achieve the mapping from incomplete EDM (2D EDM with occluded joints) to the corresponding complete EDM, inspired by the DMR. The second one is the sparse coder. Learned from sparse representation method for 3D pose estimation \cite{Sparse3D}, the incomplete EDM could be recovered by sparse coding associated with the observed part of incomplete EDMs and an over-complete dictionary which is learned from a large set of complete EDMs provided by the Human3.6M dataset \cite{Human3.6:1,Human3.6:2}.


\begin{figure}[!t]
\centering
\includegraphics[width=3.5in]{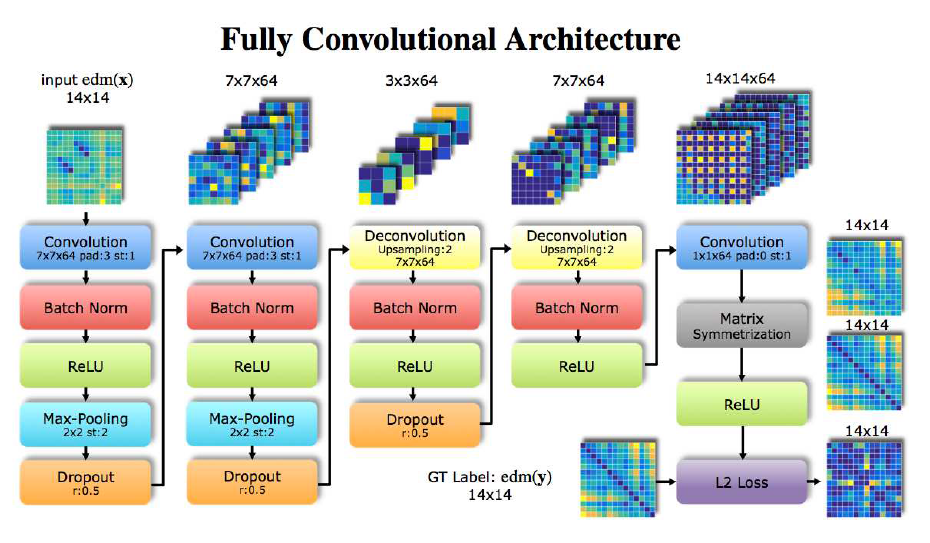}
\caption{Fully convolutional network architecture \cite{DisMatrix}, which is built upon the complete 2D measurement matrix. In this work, we use this framework as a baseline and extend the framework to handle occluded 2D joints.}
\label{fig_FConv}
\end{figure}

\section{Baseline}
We choose DMR \cite{DisMatrix} as our baseline to evaluate the performance of our 2D joints recovery methods. DMR is a recent state-of-the-art 3D human pose estimation method which adopts the two-step pipeline (see Fig.~\ref{fig_FConv}). In this paper, we are more interested in the distance matrix regression from 2D to 3D.

\subsection{DMR Step Two}
DMR uses two types of neural networks to achieve 2D-to-3D regression. DMR represents 2D and 3D poses in EDM space rather than Cartesian space.

\subsubsection{EDM}
For a pose $\mathbf{P}$ consisting of $N \times 2$ or $N\times 3$ vectors (the coordinates of $N$ joints in 2D or 3D in Cartesian space), the corresponding $N\times N$ EDM $\mathbf{E}$ is calculated as:
\begin{equation}
\mathbf{E}(i,j) = \|  p_i - p_j \|_2, ~~p_i,p_j \in \mathbf{P}.
\end{equation}

\begin{table}[!t]
\renewcommand{\arraystretch}{1.3}
\caption{Results of Experiment on EDM Application}
\label{table_1}
\centering
\begin{tabular}{|p{3cm}|p{1cm}|p{1cm}|p{1cm}|p{1cm}|}
\hline
Num. Missing Joints                       & 0 & 1 & 2 & 3\\
\hline
ave($Err_{ave}$) of FConv                 & 9.27 & 21.26 & 22.50 & 23.74\\
\hline
ave($Err_{ave}$) of Retrained FConv       & 17.40* & 17.16 & 17.25 & 17.25\\
\hline
\multicolumn{5}{l}{Retrained FConv is the above FConv which is retrained on the dataset} \\
\multicolumn{5}{l}{with corresponding number of joints missed}\\
\multicolumn{5}{l}{*This is the error of Retrained FConv tested on no missing joint dataset}
\end{tabular}
\end{table}

\subsubsection{Neural Network Regressors}
DMR adopts two types of neural networks. The first one is a fully connected network (FConn) \cite{DisMatrix} which includes three fully connected layers. The second one is a fully convolutional network (FConv) \cite{DisMatrix}, which is more robust to occluded joints and is the foundation of our first method. Fig.~\ref{fig_FConv} is obtained from the original paper \cite{DisMatrix}, which shows the structure of FConv. We could see both the input EDM and output EDM are $N\times N$ matrix ($N = 14$ in this case). The architecture consists of two convolution layers and two deconvolution layers followed by one more convolution layer at the end to compress 64 matrices into one.

\subsection{Implementation and Experiment}
During our implementation of FConv, we do not apply the matrix symmetrization layer to enforce the symmetrization, as we would like to see whether the network could learn this feature. We use caffe \cite{Caffe}, which is a brilliant deep learning framework, to implement the architecture and training with the same training settings in the original paper of DMR \cite{DisMatrix}.

In this paper, we define the performance metric of network learning  as follow:
\begin{equation}
\mathbf{Err} = |\mathbf{E_{est}} - \mathbf{E_{gt}}| / \mathbf{E_{gt}} \times 100,
\end{equation}
where $\mathbf{E_{est}}$ is the estimated 3D pose EDM and $\mathbf{E_{gt}}$ is the ground truth of corresponding 3D pose EDM. $\mathbf{Err}$ is an $N\times N$ matrix, which represents the percentage of difference between each element in estimation and ground truth matrix. Then,
\begin{equation}
Err_{ave} = \sum\limits_{i\neq j}^N {\mathbf{Err}(i,j)}/(N^2-N),
\end{equation}
where $Err_{ave}$ is the average error of one pose without considering the diagonal of the matrix, as the values on diagonal do not contribute to pose estimation and do not affect the accuracy. (3) is the main error measure we use in this paper.

To perform the experiment on the baseline we build, we choose to use the Human3.6M dataset \cite{Human3.6:1,Human3.6:2}. Human3.6M provides a subset, called Human80K, which has a more straightforward internal organization. Human80K provides 2D joints positions in image pixels and corresponding 3D joints positions in the real world (mm) covering 15 action categories. We divide 5/6 of the data as training data and the rest as testing data. In Table \ref{table_1}, the second row shows the testing results of the network trained with EDMs with no occluded joints on testing sets with from zero to three occluded joints correspondingly. As we can see, the regressor performs well when no joints are missing. However, the error shows an obvious increase when missing joints exist and increase significantly as the number of missing joints increase. These observations motivate our approaches in The weakness to occluded observations brings the demand for missing joints recovery methods.

\begin{figure}[!t]
\centering
\includegraphics[width=3.6in]{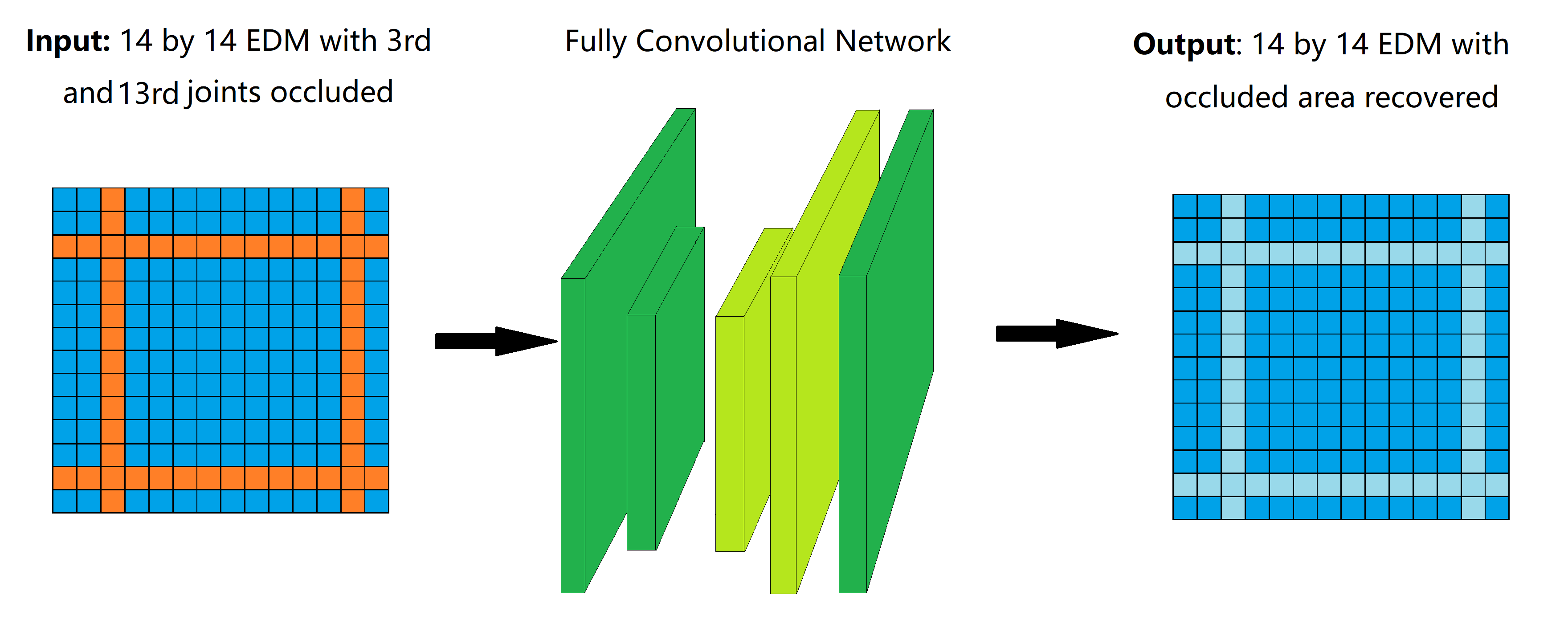}
\caption{A flow chart of our FConv based 2D joints recovery approach, which regresses between occluded EDM and complete EDM.}
\label{fig_workflowFC}
\end{figure}

\begin{figure*}[!t]
\centering
\includegraphics[width=6.2in]{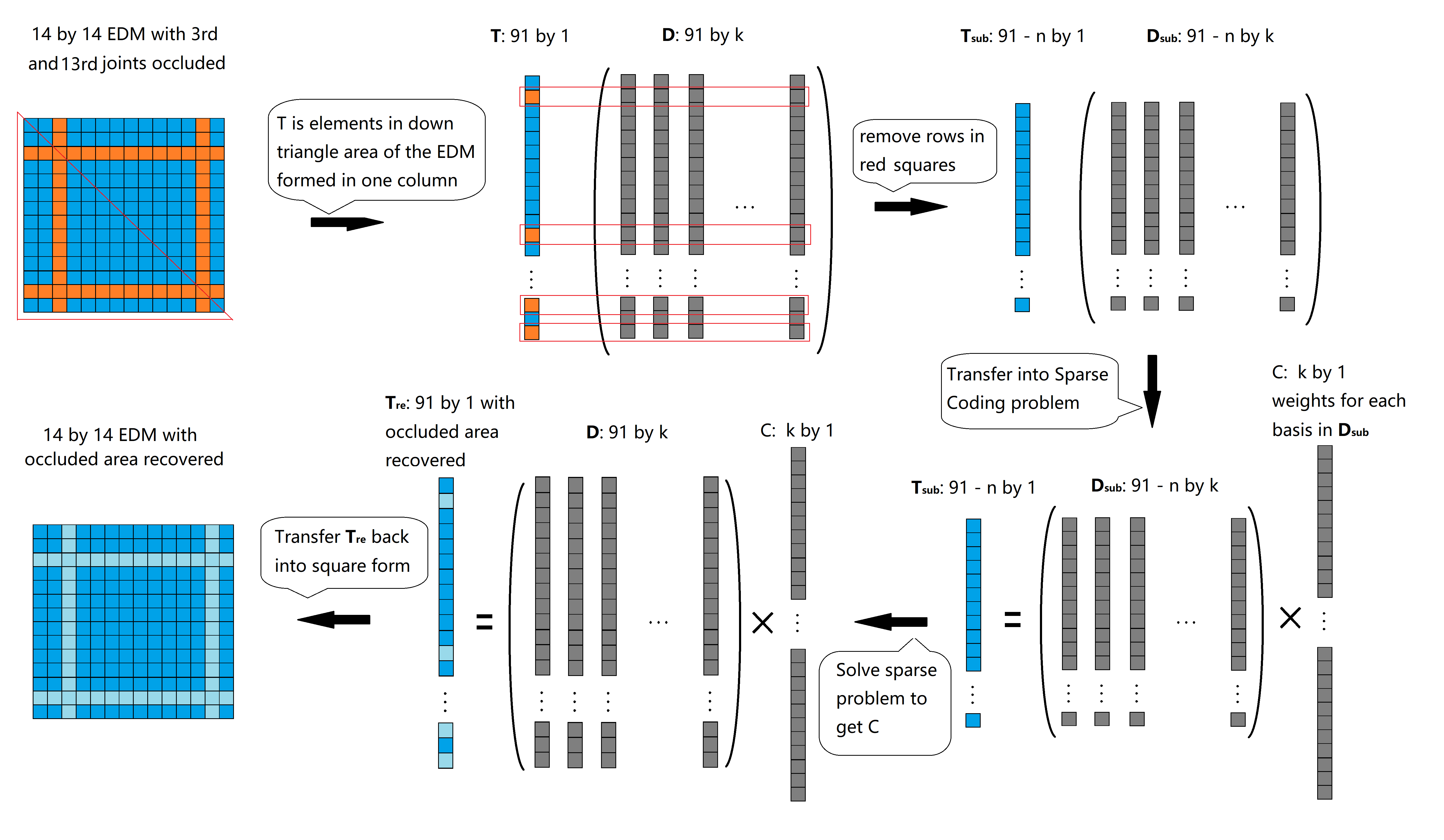}
\caption{A flow chart of our sparse coding based 2D joint recovery approach, which is built upon dictionary learning. We first solve a sparse coding problem to localize the sparse poses from which we recover the complete pose.}
\label{fig_workflowSC}
\end{figure*}

\section{Our Approaches}
\subsection{Scaling and Representation of Occluded Joints}
In order to benefit from EDM'\,s scaling invariance feature, we need to normalize the 2D poses by transforming the coordinates of 2D joints into [-1 1]. However, this would naturally bring a problem of how to normalize poses partial joints are occluded. To resolve the difficulty, we come up with two representation in dealing with the occluded joints.

The first one is called `zero'\,. Assuming the set of all observed joints of a pose is $\mathbf{P_{ob}}$, $\mathbf{P_{ob}}$ is a subset of the whole pose set $\mathbf{P}$. Let the number of observed joints be $n$, we first normalize the set $\mathbf{P_{ob}}$, and get the corresponding $n \times n$ $EDM_{ob}$ using (1). For $p_i,p_j \in \mathbf{P}$ and $p_x,p_y \in \mathbf{P_{ob}}$, the full size $N\times N$ $EDM$ of pose $\mathbf{P}$ is calculated as follows:
\begin{equation}
EDM(i,j) = \left\{ \begin{array}{rl}
 EDM_{ob}(x,y) &\mbox{ if $p_i ~is ~p_x ~and ~p_j ~is ~p_y$} \\
  0 &\mbox{ otherwise}
       \end{array} \right.
\end{equation}

The second process is `average'\,. We fill the coordinates of occluded joints with the average values of the coordinates of all observed joints. We normalize the $N \times 2$ joints coordination set. The complete $N\times N$ EDM is calculated using (1).

The first representation method sets occluded joints¡¯ corresponding rows and columns in EDM to be zero, while the second one sets average values to occluded joints¡¯ coordinates and normalizes the joints coordination set as normal. These two representations bring different features on the final EDM. The first one sets zeros to corresponding occluded area in EDM, while the second produces a more regular EDM.

These two representations have a common drawback. With the highest and lowest joints observed, which joints are occluded would not affect the normalization results of observed joints. However, if the highest or lowest joints are missed, the normalization results of all other joints would show observed differences compared with the results when they are not occluded. This inconsistency of normalization performance brings the primary source of error to our joints recovery methods, particularly in the second one.

\begin{table*}[!t]
\renewcommand{\arraystretch}{1.3}
\caption{Results on Human80K}
\label{table_2}
\centering
\scriptsize
\begin{tabular}{|c|c|c|c|c|c|c|c|c|c|c|c|c|c|c|c|c|}
\hline
Methods                      & Direct & Discuss & ~Eat~ & Greet & Phone & Pose~ & Purch & ~Sit~ & SitD~ & Smoke & TakeP & Wait & Walk & WalkD & WalkT & Overall\\
\hline
zero\_1mis                 & 4.53 & 4.26 & 4.94 & 4.46 & 6.42 & 5.00 & 4.34 & 5.02 & 7.08 & 5.18 & 5.01 & 5.12 & 4.65 & 4.73 & 4.73 & 5.03\\
\hline
zero\_2mis                 & 8.20 & 8.70 & 10.11 & 9.51 & 10.34 & 8.56 & 8.64 & 10.71 & 13.36 & 9.62 & 9.13 & 9.00 & 9.23 & 9.70 & 9.24 & 9.60\\
\hline
zero\_3mis                 & 13.43 & 13.12 & 13.15 & 13.70 & 14.77 & 12.28 & 13.40 & 16.00 & 20.35 & 14.05 & 13.37 & 13.05 & 12.67 & 14.28& 13.26 & 13.96\\
\hline
zeroMix                 & 8.17 & 8.22 & 8.93 & 8.66 & 10.05 & 8.25 & 7.96 & 10.15 & 13.05 & 9.12 & 8.73 & 8.63 & 8.33 & 9.08 & 8.64 & 9.05\\
\hline
ave\_1mis                 & 5.50 & 5.15 & 5.57 & 5.01 & 5.73 & 5.09 & 5.05 & 5.74 & 6.86 & 5.87 & 5.91 & 6.09 & 5.50 & 5.61 & 4.99 & 5.58\\
\hline
ave\_2mis                 & 9.10 & 8.81 & 9.21 & 9.86 & 10.27 & 9.41 & 8.75 & 10.78 & 13.27 & 9.90 & 10.23 & 9.63 & 9.34 & 9.64 & 9.31 & 9.82\\
\hline
ave\_3mis                 & 14.51 & 12.76 & 13.38 & 12.57 & 15.54& 13.13 & 12.24 & 15.84 & 20.08 & 14.46 & 14.65 & 13.05 & 12.56 & 14.74 & 12.65 & 14.12\\
\hline
aveMix                 & 9.00 & 8.30 & 8.72 & 8.47 & 9.79 & 8.65 & 8.00 & 10.13 & 12.56 & 9.42 & 9.64 & 9.01 & 8.53 & 9.30 & 8.44 & 9.19\\
\hline
spar\_1mis                 & 3.63 & 3.38 & 3.75 & 3.01 & 3.86 & 3.62 & 3.24 & 4.23 & 5.66 & 4.03 & 4.13 & 3.93 & 3.83 & 4.02 & 3.48 & 3.85\\
\hline
spar\_2mis                 & 7.26 & 7.16 & 7.22 & 7.85 & 8.46 & 7.98 & 6.63 & 8.16 & 11.66 & 7.67 & 8.19& 8.00 & 7.93 & 7.87 & 7.84 & 7.98\\
\hline
spar\_3mis                 & 13.18 & 11.14 & 11.63 & 10.65 & 13.91 & 12.06 & 9.58 & 13.37 & 17.28 & 12.89 & 12.49 & 12.27 & 11.60 & 12.79 & 11.13 & 12.38\\
\hline
sparMix                & 8.02 & 7.23 & 7.54 & 7.17 & 8.74 & 7.89 & 6.48 & 8.59 & 11.54 & 8.20 & 8.27 & 8.07 & 7.79 & 8.23 & 7.48 & 8.07\\
\hline
\hline
baslinMix                 & 22.75 & 22.09 & 22.37 & 22.80 & 22.16 & 22.04 & 22.82 & 23.31 & 23.97 & 22.50 & 22.85 & 22.23 & 21.65 & 22.46 & 21.69 & 22.50\\
\hline
applyzero                 & 7.80 & 8.22 & 9.18 & 8.39 & 10.38 & 9.19 & 9.18 & 13.95 & 17.65 & 10.60 & 9.56 & 9.85 & 9.08 & 9.81 & 8.08 & 10.03\\
\hline
applyave                 & 7.90 & 8.23 & 9.21 & 8.43 & 10.50 & 9.23
& 9.29 & 13.93 & 17.60 & 10.60 & 9.76 & 9.98 & 9.06 & 9.83 & 8.12 & 10.08\\
\hline
applyspar                & 7.46 & 7.91 & 8.71 & 8.05 & 9.94 & 8.95 & 8.86 & 13.32 & 17.43 & 10.17 & 9.20 & 9.81 & 8.89 & 9.57 & 7.75 & 9.70\\
\hline
zero+basl & 7.95 & 8.37 & 8.81 & 8.31 & 10.11 & 9.25 & 8.87 & 13.41 & 16.46 & 10.61 & 9.44 & 10.29 & 9.05 & 9.62 & 7.85 & 9.86 \\
\hline
ave+basl & 8.00 & 8.28 & 8.83 & 8.31 & 10.27 & 9.28 & 9.01 & 13.56 & 16.38 & 10.69 & 9.59 & 10.40 & 8.99 & 9.60 & 7.84 & 9.90 \\
\hline
\end{tabular}
\end{table*}

\subsection{FConv}
The DMR attempts to increase FConv's robustness to joints occlusion by retraining their FConv network using incomplete EDMs with random rows and columns set to zero. We retrained and tested our application on datasets with one to three joints missed separately and Table \ref{table_1} shows the results. We could see some improvement of this method compared with the performance of the previous network, but there is still a considerable potential for improvement. Also, we could see retrained method decreases the performance of network significantly on testing data with no occluded joints.

Retrain method proves that FConv has the potential to recover the occluded joints in 2D space. We believe that the shallow layers and relatively simple structure of FConv limit its ability in handling 2D-to-3D regression and joints recovery at the same time. Thus, we propose to use this network architecture to recover occluded 2D joints. We randomly pick one, two and three joints of each pose in 2D to be occluded joints. The methods use incomplete 2D joints EDMs as data and corresponding complete 2D joints EDMs as label to form the training and testing dataset. As the output is also a single $N\times N$ matrix, we could use the same FConv structure as we applied in the baseline. In this way, FConv will take a 2D EDM $\mathbf{E}$ with the occluded area as input and output a 2D EDM $\mathbf{E_{re}}$ with corresponding recovered area. The work flow of this method is shown in Fig.~\ref{fig_workflowFC}.

\subsection{Sparse Coding based Solution}
\subsubsection{Sparse Representation of EDM}
Because EDM is symmetric, and all the diagonal values zeros, we only need $N(N - 1) / 2 = 91 (N = 14)$ elements to represent an $N \times N$ EDM in order to save computation cost and storage space. Inspired from a model of sparse representation for 3D shape estimation \cite{Sparse3D,SparseMD}, every 91 dimensional vector $\mathbf{T}$ could be represented as a linear combination of each basis $\mathbf{B}$ of a pre-learned over-complete dictionary $\mathbf{D}$:
\begin{equation}
\mathbf{T} = \sum\limits_{i=1}^k {c_i \mathbf{B_i}},
\end{equation}
where $\mathbf{T}$ is the transformed version of a EDM,  $\mathbf{B_i}$ is one of the basis $91\times1$ vector in $\mathbf{D}$ and $c_i$ is the corresponding weight. This sparse representation owns the ability to model large variates of human pose, which is EDMs in this case \cite{Posecond,SparseMD,Sparse3D}. The over-complete dictionary $\mathbf{D}$, which is $\{ \mathbf{B_1}\mathbf{B_2}...\mathbf{B_k}\}$, is learned from a large motion capture dataset to enable encoding features of enormous EDMs.

\subsubsection{Occluded Joints Recovery}
As $\mathbf{D}$ is over-complete, the unobserved elements of $\mathbf{T}$ could be recovered by inferring observed parts with basic vectors $\mathbf{B}$. This formulates the recovery problem into a sparse coding problem. For retrieving unobserved elements set $\mathbf{M}$ in $\mathbf{T}$, we get:
\begin{equation}
\mathbf{T_{sub}} = \mathbf{T} - \mathbf{M}, ~\mathbf{B_{subi}}(x) = ~\mathbf{B_i}(j), ~if ~\mathbf{T_{sub}}(x) ~is~ \mathbf{T}(j),
\end{equation}
where $1 \leq i \leq k,~1 \leq j \leq 91 ~and ~ \mathbf{D_{sub}} = \{ \mathbf{B_{sub1}}\mathbf{B_{sub2}}...\mathbf{B_{subk}}\}$. Now, we get a occluded joints unrelated sparse coding problem \{ $\mathbf{T_{sub}}$, $\mathbf{D_{sub}}$, $\mathbf{c}$ \}, which requires to find $\mathbf{c}$ which minimizes the following function:
\begin{equation}
\min \{ \| \mathbf{T_{sub}}-\mathbf{D_{sub}} \mathbf{c} \|^2 + \lambda \|\mathbf{c}\|_1 \},
\end{equation}
where $\lambda$ is the penalty term which enforces the sparsity of $\mathbf{c}$. We introduce a sparse coding solution which is part of Kernel version of the sparse representation classifier application which is implemented with CVX \cite{AnalyticEst,FaceRec,CVX1,CVX2} to solve the problem (7). After we get $C$, then could get recovered $\mathbf{T_{re}}$ is calculated as follow:
\begin{equation}
\mathbf{T_{re}} = \mathbf{D}\mathbf{c},
\end{equation}
The demonstration of this method is in Fig.~\ref{fig_workflowSC} for easier understanding.

\subsubsection{Dictionary Learning}
As mentioned above, the dictionary we used should be big enough so that it could model large variates of EDMs and enable good recovery performance. In this case, dictionary learning techniques are necessary to learn a dictionary out of big training dataset. Basically, the dictionary learning problem is formulated as follows:
\begin{equation}
\mathbf{argmin_{D,\alpha}} 1/2 \{   \| \mathbf{S}-\mathbf{D} \alpha\|_2^2 + \lambda \|\alpha\|_1 \},
\end{equation}
where $\mathbf{S}$ is the training dataset, $\alpha$ is the corresponding weights set. The connection between (9) and (7) is that $\mathbf{S}$ and $\alpha$ consist of multiple $\mathbf{T_{sub}}$s and $\mathbf{c}$s respectively. The difference is (9) has two variables while (7) only has one.

To solve the problem, we choose the method that updates $\alpha$ and $\mathbf{D}$ alternatively. When $\mathbf{D}$ is fixed, $\mathbf{S}$ and $\alpha$ could be separated into multiple individual samples $\mathbf{T}$s and corresponding $\mathbf{C}$s. In this way, the problem is divided into multiple sparse coding problems which could be solved by (7). When $\alpha$ is fixed, this becomes a least-squares problem, and lots of solution are developed to solve it, like the K-SVD \cite{DictionModel}. However, these methods become time-consuming when dictionary and training set are too large. Thus, we introduce an on-line dictionary learning method to update $\mathbf{D}$ at this stage based on stochastic approximations \cite{onlineDL} with high efficiency in large training sets. The Matlab code of this algorithm we use is from \cite{Code:onlineDL1,Code:onlineDL2}.

\section{Experimental Results}
As we concern the recovery results of occluded joints, there is one more step that we could form the final recovered EDM $\mathbf{E_{f}}$ from $\mathbf{E_{re}}$ and $\mathbf{T_{re}}$
\begin{equation}
\mathbf{E_f}(i,j) = \left\{ \begin{array}{rl}
  \mathbf{E}(i,j)  &\mbox{ if $p_i,p_j \in \mathbf{P_{ob}}$} \\
  \mathbf{E_{re}}(i,j)       &\mbox{ otherwise}
       \end{array} \right.
\end{equation}
where $1 \leq i \leq 14,~1 \leq j \leq 14$, $\mathbf{E}$ and $\mathbf{E_{re}}$ are the input and output of our recovery method respectively and $\mathbf{P_{ob}}$ is the set of observed joints for pose set $\mathbf{P}$. In terms of $\mathbf{T_{re}}$, we need to reform this $91\times 1$ vector back into $14\times14$ matrix $\mathbf{E_{re}}$ and then use (10). Equation (10) tells we only need the values in corresponding rows and columns of occluded joints in $\mathbf{E_{re}}$ and assign them back to the original input $\mathbf{E}$. In a perfect world, we do not need (10) because the recovery method is designed to find a full matrix from its knowledge that fit the values in observed parts of input matrix perfectly. But in practice, it cannot always find the exact fit. Hence, we need to use Eq-(10) to extract the values of recovered occluded parts in $\mathbf{E_{re}}$ without introducing unnecessary error. To measure the performance of our recovery methods, we use (3) and (2) by simply replacing the $\mathbf{E_{est}}$ with $\mathbf{E_f}$.

\begin{figure}[!t]
\centering
\includegraphics[width=2.8in, height=1.8in]{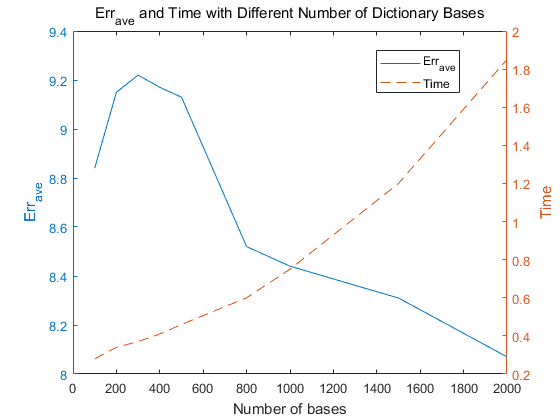}
\caption{$Err_{ave}$ and time of sparse coding with different number of dictionary bases.}
\label{Dictionarybases}
\end{figure}

\begin{figure}[!t]
\centering
\includegraphics[width=2.8in, height=1.8in]{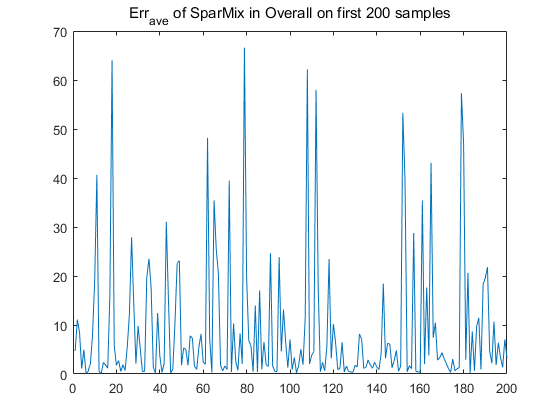}
\caption{$Err_{ave}$ of SparMix in overall on first 200 samples.}
\label{SparMix}
\end{figure}

As mentioned in baseline, we decide to use Human80K, a subset of Human3.6M \cite{Human3.6:1,Human3.6:2}, to conduct experiments on our occluded joints recovery methods. Human80K provides over 50,000 samples covering 15 action categories. To ensure the consistency with our baseline, we use the same division of training and testing dataset to perform training and testing independently. As the dataset is relatively small, training is done over all action categories, while testing is carried out in each category separately.

Human80K provides 2D human pose with 2D joints coordinates in image pixels. For each set of 2D joints $\mathbf{P}$, we randomly pick one, two and three joints to be occluded independently to form three different 2D joints sets. Along with original one, we get four different 2D joints sets from one pose. Then we use our two types of representations to normalize and form the corresponding EDMs. For each type of normalization method, we get three occluded EDMs, which are from three occluded joints sets, and one complete EDM, which is from the original joints set.

\subsection{FConv}
We did experiments on one to three occluded joints EDMs separately and then mixed them together to do an overall experiment. As the values of the 14$\times$14 matrix are all fed into the network, we did the experiment of our network on two datasets which are generated by two occluded joints representation methods to compare the performance of these two occluded joints representation methods. Then we feed the recovery results to our baseline and measure the performance. Finally, we connect our network which is trained on mixed missing joints dataset and the baseline network in sequel. Then we use mixed missing joints EDMs as data and corresponding 3D EDMs as label to retain the connected network to see if there are any further improvement.

\subsection{Sparse Coding}
Since the corresponding values of occluded joints in $\mathbf{T}$ is not going into our sparse coding algorithm, we choose the `average'\, as the type of representation method would not affect the performance of the algorithm. For this method, the number of basis vector in the dictionary would be tricky as it directly affects the ability to recover occluded parts in EDMs. We test from 100 to 2000 bases of the dictionary and test the performance of corresponding sparse coding methods on mixed occluded joints dataset. As shown in Fig.~\ref{Dictionarybases}, we decide to use 2000 in the following experiment with the consideration of accuracy and computation time consumption. It is worth noting that, rather than FConv, we only train our dictionary on complete EDMs as we hope the dictionary encode the features of human poses with complete joint sets.

\subsection{Comparison of Performance}
Overall, from Table \ref{table_2}, we could see our two approached do outstanding works in recovering occluded EDMs. With the increase of the number of occluded joints, it is more difficult to recover the EDMs. Also, we could figure out some of the categories, like 'Sit', 'SitD' and 'Smoke', are always harder than others for our methods. After we applied our methods in the baseline ('applyzero', 'applyave' and 'applyspar'), we could see the significant improvement of performance of the system under conditions of one to three random joints occluded. Also, the result of our retrained network with the connection of FConv recovery net and baseline ('zero+basl' and 'ave+basl') further improved in most categories compared with the result of simple applications ('applyzero' and 'applyave').

For comparison, we could see that the results of sparse coding are better than FConv in all categories for both recover measurement and performance in pipeline applications. However, sparse coding is much more unstable than FConv. The variance of sparse coding of 'sparMix' on 'Overall' is 190.98, while the variance of 'zeroMix' on 'Overall' is only 119.85. Fig.~\ref{SparMix} shows sparse coding could precisely recover the occluded joints in most time, but it could also be far from the ground truth. When we look further into the samples, the large error of sparse coding always happen when the drawback of representation that we mentioned before occur. On the other hand, FConv'\,s performance is much stable. Even the error of FConv also raises when the weakness of representation occur, but the increase would be relatively smaller. Compared with `zero'\, occluded joints representation methods, `average'\,is little less competitive in overall performance, but it tends to perform better in the 'hard' categories.

\textbf{Running Time}
For computation time, with the benefits of advanced deep learning techniques in both software and hardware and differences in computation complexity, FConv approach is processed much faster on GPU (NVIDIA GeForce GTX 1070) with average 0.00018s per sample while sparse coding on CPU (Intel(R) Core(TM) i7-6700K at 4.00GHz) needs  1.85s per sample on average.

\section{Conclusion}
In this paper, we proposed two effective approaches for occluded joints recovery in 3D human pose estimation based on the Euclidean Distance Matrix. The first approach uses deep neural networks to enable the regression between occluded and complete distance matrix. The second approach uses a learned over-complete dictionary as knowledge to achieve much more precise results. Both approaches achieve super performance over state-of-the-art methods. In future, we would like to exploit the connection between these two approaches and integration of both approaches under a unified framework. Furthermore, extending the current single image based approach to videos \cite{NRSFM_CVPR_2012,NRSFM_ICCV_2017} is another interesting direction.


\section*{Acknowledgment}
This project was supported in part by National 1000 Young Talents Plan of China, Natural Science Foundation of China (61420106007, 61671387), and ARC grant (DE140100180).




\begin{thebibliography}{1}
\bibitem{Posecond}
I. Akhter and M. J. Black. Pose-conditioned joint angle limits for 3D human pose reconstruction, in \emph{IEEE Conference on Computer Vision and Pattern Recognition}, 2015, pp. 1446-1455.

\bibitem{CVX1}
M. Grant; S. Boyd, CVX: Matlab Software for Disciplined Convex Programming, version 2.1, \emph{http://cvxr.com/cvx}, 2014.

\bibitem{CVX2}
M. Grant; S. Boyd, Graph implementations for nonsmooth convex programs, Recent Advances in Learning and Control, \emph{Springer-Verlag Limited}, 2008, pp.95-110.

\bibitem{AnalyticEst}
B. Gaonkar; C. Davatzikos, Analytic estimation of statistical significance maps for support vector machine based multi-variate image analysis and classification, in \emph{NeuroImage}, 2013, 78: 270-283.

\bibitem{Code:onlineDL1}
J. Greggson, Matlab implementation of online dictionary learning with example driver code, [Source code], https://github.com/jamesgregson/matlab\_dictionary\_learning

\bibitem{Code:onlineDL2}
J. Greggson, Basic Dictionary Learning in Matlab, [Source code], http://www.jamesgregson.ca/uncategorized/basic-dictionary-learning-in-matlab/

\bibitem{relatedwork:oimage}
J. B. Huang and M. H. Yang. Estimating Human Pose from Occluded Images, in \emph{Asian Conference on Computer Vision}, 2009, pp 48-60.

\bibitem{Human3.6:1}
C. Ionescu; D. Papava; V. Olaru; C. Sminchisescu, Human3.6M: Large Scale Datasets and Predictive Methods for 3D Human Sensing in Natural Environments, in \emph{IEEE Transactions on Pattern Analysis and Machine Intelligence}, 2014, 36(7):1325-1339.

\bibitem{Human3.6:2}
C. Ionescu, F. Li and C. Sminchisescu. Latent Structured Models for Human Pose Estimation, in \emph{International Conference on Computer Vision}, 2011, pp. 2220-2227.

\bibitem{Caffe}
Y. Jia; E. Shelhamer; J. Donahue; S. Karayev; J. Long; R. Girshick; S. Guadarrama; T. Darrell, Caffe: Convolutional Architecture for Fast Feature Embedding, in \emph{arXiv preprint arXiv:1408.5093}, 2014, abs/1408.5093: 1-4.

\bibitem{DisMatrix}
F. Moreno-Noguer. 3D Human Pose Estimation from a Single Image via Distance Matrix Regression, in \emph{IEEE Conference on Computer Vision and Pattern Recognition}, 2017, pp. 2823-2832.

\bibitem{onlineDL}
J. Mairal, F. Bach, J. Ponce, and G. Sapiro. Online dictionary learning for sparse coding, in \emph{International Conference on Machine Learning}, 2009, pp. 689-696.

\bibitem{relatedwork:selfoc}
I. Radwan, A. Dhall, and R. Goecke. Monocular image 3D human pose estimation under self-occlusion. in \emph{International Conference on Computer Vision}, 2013, pp. 1888-1895.

\bibitem{NRSFM_CVPR_2012}
Y. Dai, H. Li and M. He. A simple prior-free method for non-rigid structure-from-motion factorization, in \emph{IEEE Conference on Computer Vision and Computer Vision}, 2012, pp. 2018-2025.

\bibitem{NRSFM_ICCV_2017}
S. Kumar, Y. Dai and H. Li. Monocular Dense 3D Reconstruction of a Complex Dynamic Scene from Two Perspective Frames, in \emph {International Conference on Computer Vision}, 2017, pp. 4649-4657.

\bibitem{relatedwork:occldete}
I. Radwan, A. Dhall, J. Joshi, and R Goecke. Regression Based Pose Estimation with Automatic Occlusion Detection and Rectification, in \emph{IEEE International Conference on Multimedia and Expo}, 2012, pp.121-127.

\bibitem{DictionModel}
R. Rubinstein; A. M. Bruckstein; M. Elad, Dictionaries for Sparse Representation Modeling, in \emph{Proceedings of the IEEE}, 2010, 98(6): 1045-1057.

\bibitem{BayesianBased}
M. Sanzari, V. Ntouskos and F. Pirri, Bayesian Image Based 3D Pose Estimation, in \emph{European Conference on Computer Vision}, 2016, pp. 566-582.

\bibitem{FaceRec}
J. Wright; A. Y. Yang; A. Ganesh; S. S. Sastry; Y. Ma, Robust Face Recognition via Sparse Representation, in \emph{IEEE Transactions on Pattern Analysis and Machine Intelligence}, 2009, 31(2): 210 - 227.

\bibitem{DualSource}
H. Yasin, U. Iqbal, B. Kruger, A. Weber, and J. Gall. A Dual-Source Approach for 3D Pose Estimation from a Single Image. in \emph{IEEE Conference on Computer Vision and Pattern Recognition}, 2016, pp. 4948-4956.

\bibitem{Sparse3D}
X. Zhou; M. Zhu; S. Leonardos; K. Daniilidis, Sparse Representation for 3D Shape Estimation: A Convex Relaxation Approach, in \emph{IEEE Transactions on Pattern Analysis and Machine Intelligence}, 2017, 39(8): 1648-1661.

\bibitem{SparseMD}
X. Zhou, M. Zhu, S. Leonardos, K. G. Derpanis and K. Daniilidis. Sparseness Meets Deepness: 3D Human Pose Estimation from Monocular Video, in \emph{IEEE Conference on Computer Vision and Pattern Recognition}, 2016, pp. 4966-4975.


\end{thebibliography}
%
\balance

\end{document}